\documentclass[letterpaper, 10 pt, conference]{ieeeconf}  

\IEEEoverridecommandlockouts                              

\usepackage{times}
\usepackage[ruled,vlined]{algorithm2e}
\usepackage{multicol}
\usepackage[bookmarks=true]{hyperref}
\usepackage{graphicx}
\usepackage{tabularx}
\usepackage{amssymb}
\usepackage{amsmath}
\usepackage{fancyhdr}
\usepackage{algorithmic}
\usepackage{tikz}
\usepackage{bm}
\usepackage{multicol}
\usepackage{listings}
\usepackage{subfigure}

\algsetup{linenosize=\small}
\newcommand{\vect}[1]{\mathbf{#1}} 
\newcommand{\vectg}[1]{\boldsymbol{#1}}

\newcommand{\norm}[1]{{\|{#1}\|}}
\def\NAT@parse{\typeout{IEEEtran error: Attempt to use fake Natbib command 
which is provided to fool Hyperref.}}

\begin{document}

\title{\LARGE \bf Post-Stall Navigation with Fixed-Wing UAVs using Onboard Vision}

\author{Adam Polevoy$^{1}$, Max Basescu$^{1}$, Luca Scheuer$^{1}$, Joseph Moore$^{1}$
\thanks{$^{1}$Johns Hopkins University Applied Physics Lab, Laurel, MD, 20723
        {\tt\small \{Adam.Polevoy,Max.Basescu,Luca.Scheuer, Joseph.Moore\}@jhuapl.edu}{\newline
        DISTRIBUTION STATEMENT A: Approved for public release; distribution is unlimited.}}
        }


%

\maketitle

\begin{abstract}

Recent research has enabled fixed-wing unmanned aerial vehicles (UAVs) to maneuver in constrained spaces through the use of direct nonlinear model predictive control (NMPC) \cite{basescu2020direct}. However, this approach has been limited to a priori known maps and ground truth state measurements. In this paper, we present a direct NMPC approach that leverages NanoMap \cite{florence2018nanomap}, a light-weight point-cloud mapping framework to generate collision-free trajectories using onboard stereo vision. We first explore our approach in simulation and demonstrate that our algorithm is sufficient to enable vision-based navigation in urban environments. We then demonstrate our approach in hardware using a 42-inch fixed-wing UAV and show that our motion planning algorithm is capable of navigating around a building using a minimalistic set of goal-points. We also show that storing a point-cloud history is important for navigating these types of constrained environments. 

\end{abstract}

\IEEEpeerreviewmaketitle

\section{Introduction}


Rotary-wing unmanned aerial vehicles (UAVs) are often preferred for navigating in constrained environments. Not only are they capable of near zero turn radii, but many are amendable to differentially flat representations that can greatly reduce computational requirements for online trajectory optimization \cite{mellinger2011minimum}. By contrast, fixed-wing UAVs offer significant advantages over rotary-wing UAVs in terms of energy efficiency, endurance, and speed. Traditionally fixed-wing UAVs have been hindered by limited maneuverability and restricted to large open spaces. Recent research has shown that aerobatic fixed-wing UAVs are capable of navigating through complex environments by exploiting the full flight envelope \cite{levin2017agile,basescu2020direct}. By operating beyond the conventional flight regimes, there is an opportunity to achieve both range and maneuverability in a single vehicle.

In \cite{basescu2020direct}, the authors demonstrated that a fixed-wing UAV could navigate a narrow corridor by executing post-stall turns via nonlinear model predictive control (NMPC). While this approach proved a promising means to increase fixed-wing maneuverability, it required an a priori map of the environment. For real-world applications, full a priori map information is often not practical. Rather, the vehicle must navigate in an unknown environment and construct a map during runtime for collision avoidance and navigation. 

Real-time mapping poses a number of significant challenges, for aerobatic, fast-moving fixed-wing UAVs. Slower-moving multi-rotor UAVs have long relied on sensors that leverage LIDAR (e.g.,\cite{grzonka2009towards}). However, these sensors are often too heavy for small aerobatic fixed-wings, have update rates too slow for rapid maneuvering, have returns restricted to two-dimensional spaces, or suffer from a significantly sparse set of range measurements. With regards to mapping algorithms, commonly used occupancy-grid based techniques such as \cite{hornung2013octomap} have proved to be too slow and computationally intensive for use in fast flight. These challenges are further exacerbated by state and range-sensor uncertainty, which can dramatically impact map accuracy. 


In this paper, we present an approach that leverages direct NMPC and an existing mapping framework, NanoMap \cite{florence2018nanomap}, to enable a vision-based fixed-wing navigation system capable of leveraging post-stall flight for aggressive maneuvers. Our receding-horizon controller continually plans into unknown space, using distance information from the NanoMap framework in-the-loop for both nominal low-fidelity global path generation and higher-fidelity real-time trajectory optimization. The use of NanoMap allows us to reason about sensor uncertainty as well as sensor measurement history. We demonstrate our approach in hardware using a 42-inch wingspan UAV with onboard sensing and computation. 

\begin{figure}
    \centering
    \begin{minipage}{.48\textwidth}
        \centering
        \includegraphics[trim={25 25 0 0},clip,width=0.70\linewidth]{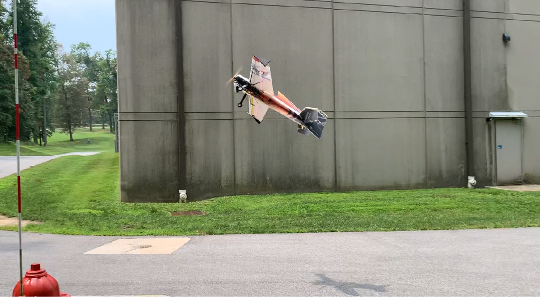}
    \end{minipage}
    \begin{minipage}{0.48\textwidth}
        \centering
        \includegraphics[trim={0 2 0 2},clip,width=0.70\linewidth]{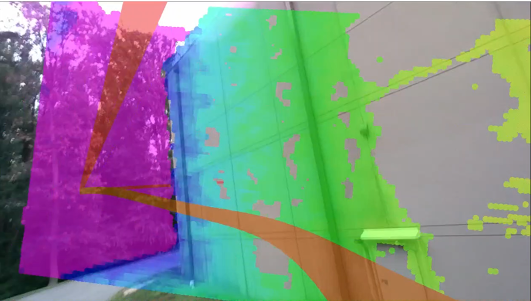}
    \end{minipage}
    \label{Fig:stallturn}
    \caption{Screenshot of aerobatic fixed-wing executing a post-stall maneuver (top) and corresponding point cloud data (bottom). The global path plan is shown in red and the current trajectory is shown in orange.}
\end{figure}

\section{Related Work}

A majority of the research into UAV navigation and control using onboard sensing has been conducted with multirotor UAVs. Early research in this area focused on LIDAR-based simultaneous localization and mapping (SLAM), and leveraging generated maps for navigation and collision avoidance \cite{bachrach2011range,shen2011autonomous,grzonka2009towards}. As the flight speeds increased, onboard vision emerged as another primary sensing solution \cite{blosch2010vision,faessler2016autonomous,huang2017visual}. Researchers also began to address the challenges associated with the computational costs of online mapping, as well as the challenges posed by large amounts of state uncertainty. \cite{tabib2021simultaneous} utilizes Gaussian Mixture Model (GMM) based mapping to reduce the memory requirements of storing and transmitting mapping data.  \cite{gao2019flying} addresses the mapping challenge by maintaining a KD-tree representation of LIDAR measurements, generating a valid flight corridor, and optimizing trajectories within that corridor. Similarly, \cite{florence2020integrated} utilizes a trajectory library approach on a KD-tree representation of current depth camera measurements to minimize the probability of collision. The authors then extend this work with a lightweight mapping data structure, NanoMap \cite{florence2018nanomap}, which maintains a history of sensor measurements. For control and planning, most of approaches cited above exploit differential flatness to generate multirotor trajectories in real-time. More recent work has begun to explore NMPC for quadcopter flight, especially in the context of perception-aware control \cite{falanga2018pampc}.

Several efforts have explored the use of onboard sensing for fixed-wing navigation. In \cite{bry2015aggressive}, the authors demonstrated the ability to fly in a constrained-space using a scanning LIDAR. Their work focused on the state-estimation problem and used a differentially flat model valid for a reduced flight envelope. In \cite{barry2018pushbroomstereo}, authors demonstrated fixed-wing flight using onboard stereo vision and a trajectory-library approach limited to conventional angle-of-attack regimes. 

A number of researchers have also focused solely on the fixed-wing planning and control problem, assuming external sensing and known environments. These efforts have placed a greater emphasis on maneuvering across unconventional attitudes and flow regimes. Early work \cite{cory2008experiments} and \cite{sobolic2009nonlinear} explored post-stall flight as well as transitions to-and-from a prop-hang configuration. In \cite{moore2014robust}, a library of trajectories was used to enable robust post-stall perching performance. Motion planning for fixed-wing UAVs was also explored in \cite{majumdar2017funnel}, where a library of funnels was used to navigate through a known obstacle field in real-time. \cite{alturbeh2014real} explored real-time trajectory motion-planning of fixed-wing UAVs using simplified models restricted to low angle-of-attack domains. In \cite{levin2017agile, khan2016modeling, bulka2018autonomous,levin2019real, bulka2019high}, researchers have developed motion planners for aerobatic fixed-wing UAVs using trajectory libraries generated offline using high fidelity physics models. In some cases, especially in the presence of high winds, NMPC has been utilized to control fixed-wing UAVs \cite{mathisen2020nmpc}. In \cite{basescu2020direct}, a receding-horizon NMPC approach was demonstrated that enabled fixed-wing UAVs to execute post-stall maneuvers to navigate constrained environments.

Here, we build on the work done in \cite{basescu2020direct} and develop an NMPC-based navigation approach capable of running in real-time using stereo vision. Our method is distinct from other approaches in that it allows seamless exploitation of the full flight envelope for navigation via NMPC, leverages a point-cloud map to maintain a history of sensor data, and provides a means for handling uncertainty in sensor data and state estimates. Through simulation, we show in the importance of compensating for sensor uncertainty, as well as the necessity of reasoning about field-of-view history, especially in the context of aggressive post-stall turns. We demonstrate our results in hardware, and to our knowledge, this is the first demonstration of NMPC for post-stall fixed-wing flight and collision avoidance using onboard vision. We also present a framework which allows for hardware flight testing with a simulated point cloud sensor, to facilitate safe at-altitude testing in virtual environments.

\label{sec:intro}
\section{Approach}

In this paper, we expand upon the control strategy proposed in \cite{basescu2020direct}. This control strategy consists of four major stages: RRT generation and pruning, spline-based smoothing, direct trajectory optimization, and local linear feedback. We make use of the same dynamics model for a larger 42-inch wingspan vehicle.

We assume the use of a mounted depth camera to perform sensing, and light-weight mapping is performed with NanoMap. RRT paths are reused whenever possible to reduce computational load and to maintain consistency between control iterations. Horizon point selection is constrained to known regions of the map to prevent trajectory generation through unobserved obstacles. Constraints on the distance to obstacles and estimated probability of collision are both evaluated for trajectory generation.

\subsection{Mapping}

NanoMap was chosen in place of OctoMap because it has a low map maintenance cost and is less likely to propagate odometry error due to its local map structure. It is also able to propagate state uncertainty through a local history of depth measurements.

A few adjustments were made to NanoMap to better suit our control strategy. Query replies were modified to include the transformation from the query body frame to the current body frame. This transformation is necessary for calculating the obstacle constraint gradients.

NanoMap query replies include information about the field of view status of the query point. This information specifies if the query point is in free space, occluded space, or outside of the field of view of depth measurements. If the query point was not within the field of view free space of any prior depth measurements, it had no valid corresponding depth measurement. In this case, the obstacle points are pulled from the most recent depth measurement. We modified NanoMap to return the K-nearest obstacle points from the most recent occluded field of view instead. This yielded more accurate nearest obstacle points, especially while planning through occluded space.

\subsection{RRT Generation}

In the base controller proposed in \cite{basescu2020direct}, a rapidly-exploring random tree (RRT) was used to generate global paths to the goal for horizon point selection. The RRT was expanded until the goal point was connected to the tree, and the path from the current position to the goal position was extracted. This path was iteratively pruned, smoothed with B\'ezier curves, and parameterized in terms of time. We refer to this final smoothed path as the smoothed RRT path, and it is used to select a receding horizon point for trajectory generation.

This strategy poses a problem for planning in unknown environments; the horizon point may be in or behind an unobserved obstacle. This is undesirable since the direct trajectory optimization problem is warm started with the previously generated trajectory. If a previously generated trajectory becomes intersected by newly obtained map data, the optimizer will often be unable to find a feasible solution.

To avoid this, the RRT path is constrained to known regions of the map using a method similar to that proposed in \cite{umari2017autonomous}. Our method is outlined in Algorithm \ref{Algo:RRT}. As the tree expands, a list of frontier nodes (which exist in unknown space and connect to nodes in known space) is maintained. A node is in a known region of the map if the NanoMap query returns a field of view status indicating that the query point is in free space. If the goal point is not connected to the tree, or if the goal point is in an unknown region of the map, then the frontier node closest to the goal position is used as the goal instead.

\begin{algorithm}
    \caption{RRT($x_{init}$, $x_{goal}$, $\Delta x$, map, $K$, $\Delta t$)}

    $\tau$.init($x_{init}$), $f\gets[]$\\
    \For {$k=1$ to $K$}
    {
        $x_{rand}\gets$ RANDOM\_STATE()\\
        $x_{near}\gets$ NEAREST\_NEIGHBOR($x_{rand}$, $\tau$)\\
        $u\gets$ SELECT\_INPUT($x_{rand}$, $x_{near}$)\\
        $x_{new}\gets$ NEW\_STATE($x_{near}$,$u$,$\Delta t$)\\
        \If{PATH\_OBST\_FREE($x_{near}$,$x_{new}$, map)}
        {
            $\tau$.add\_vertex($x_{new}$)\\
            $\tau$.add\_edge($x_{near}$,$x_{new}$,$u$)\\
            \If{DISTANCE($x_{new}$, $x_{goal}$) $\leq \Delta x$}
            {
                $x_{goal}\gets x_{new}$\\
                break\\
            }
            \If{IS\_FRONTIER\_NODE()$x_{new}$, map)}
            {
                $f$.add($x_{new}$)\\
            }
        }
    }
    \If{$x_{goal}$ not in $\tau$ or IN\_UNKNOWN($x_{goal}$, map)}
    {
        $x_{goal} \gets x_f \gets x{goal}$.FIND\_CLOSEST($f$)\\
    }
    Return $\tau$, $x_{goal}$\\
    \label{Algo:RRT}
\end{algorithm}

At each control iteration, a truncated version of the prior RRT path is used to initialize the RRT. This truncation takes into consideration both the updated vehicle state and any newly detected obstacles. The beginning of the RRT path is truncated to minimize the distance between the start of the path and the vehicle state; the end is truncated to ensure that the path does not intersect with obstacles. 

Reuse of the prior path reduces RRT computation time and also provides a more gradual evolution of the RRT path. This leads to a more incremental update of the receding horizon goal point and a reduced the computational burden for the warm-started trajectory optimization routine.

\subsection{Direct Trajectory Optimization}

While the base direct trajectory optimization problem is the same as in \cite{basescu2020direct}, NanoMap queries are used to calculate the obstacle avoidance constraints instead of Octomap. 


Our optimization problem can be written as

\begin{equation}
\begin{aligned}
& \underset{\vect{x}_k, \vect{u}_k, h }{\text{min}}
& & \sum_{k=0}^{N}\left[\vect{{x_k}^T}\vect{Q}\vect{x_k}\right]+\vect{{x_{N+1}}^T}\vect{Q_f}\vect{x_{N+1}} \\
& \text{s.t.}
& & \forall k \in [0,\ldots,N] \text{ and }\\
&
& & \vect{x}_{k} - \vect{x}_{k+1} + \frac{h}{6.0}(\vect{\dot{x}}_{k} + 4 \vect{\dot{x}}_{c,k} + \vect{\dot{x}}_{k+1})= 0\\
&
& & \vect{x}_f - \vectg{\delta}_f \le \vect{x}_{N} \le \vect{x}_f + \vectg{\delta}_f\\
&
& & \vect{x}_i - \vectg{\delta}_i \le \vect{x}_{N} \le \vect{x}_i + \vectg{\delta}_i\\
&
& & \vect{x}_{min}  \le \vect{x}_{k} \le \vect{x}_{max},~~ \vect{u}_{min}  \le \vect{u}_{k} \le \vect{u}_{max}\\
&
&  & c(\vect{x}) \geq 0 \\
&
& & h_{min}  \le h \le h_{max}
\end{aligned}
\end{equation}
where 
\begin{align}
\vect{\dot{x}}_{k} &= \vect{f}(t, \vect{x}_{k}, \vect{u}_{k}),~~
\vect{\dot{x}}_{k+1} = \vect{f}(t, \vect{x}_{k+1}, \vect{u}_{k+1})\nonumber\\
\vect{u}_{c,k} &= (\vect{u}_{k} + \vect{u}_{k+1}) / 2\nonumber\\
\vect{x}_{c,k} &= (\vect{x}_{k} + \vect{x}_{k+1}) / 2 + h (\vect{\dot{x}}_{k} - \vect{\dot{x}}_{k+1}) / 8\nonumber\\
\vect{\dot{x}}_{c,k} &= \vect{f}(t, \vect{x}_{c,k}, \vect{u}_{c,k}).
\end{align}

Here, $\vect{x}$ signifies the system state, $\vect{u}$ is the control actions, and $\vect{f}(t, \vect{x}_{k+1}, \vect{u}_{k+1})$ is the dynamical model. $h$ is the time step bounded from $h_{min}=0.001s$ to $h_{max}=0.2s$, $\vectg{\delta}_f$ and $\vectg{\delta}_i$ represent the bounds on the desired final and initial states ($\vect{x}_f$, $\vect{x}_i$), respectively. $N$ is the number of knot points, and $c(\vect{x})$ represents the collision constraint. 

The first constraint that we evaluated was $c(\vect{x}) = d(\vect{x}) - r$, or the distance to obstacles, where $r$ is the obstacle radius. Given the query, $\vect{x}$, and the $K=10$ nearest neighbors $\vect{a}_i$, the distance to each nearest neighbor is calculated as $d(\vect{x})_i = \norm{\vect{x}-\vect{a}_i}$. The distance to obstacle constraint is calculated as the average distance to the nearest neighbors $d(\vect{x}) = \frac{1}{K}\sum_{i=1}^{K}\left(d(\vect{x})_i\right)$.  Similarly, the gradient for this constraint is calculated as $\nabla d(\vect{x}) = \frac{1}{K}\sum_{i=1}^{K}\left(\frac{\vect{x}-\vect{a}_i}{\norm{\vect{x}-\vect{a}_i}}\right)$

The second constraint we evaluated was the probability of collision; we replaced $d(\vect{x}) \geq r$ with $p(\vect{x}) \leq s$, where $s$ is the maximum probability of collision. Probability of collision is calculated using the formulation presented in \cite{thomas2021integrated}. Specifically, the probability between a query point and a nearest neighbor is calculated as 
\begin{align}
p_i(\vect{x}) &= \sum_{k=0}^\infty(-1)^kc_k\frac{(r+s)^{2^{\frac{n}{2}+k}}}{\Gamma(\frac{n}{2}+k+1)} \nonumber \\
c_0 &= exp\left(-\frac{1}{2}\sum_{j=1}^n{{b_i}_j}^2\right)\prod_{j=1}^n(2\lambda_j)^{\frac{1}{2}} \nonumber \\
c_k &= \frac{1}{k}\sum_{j=0}^{k-1}d_{(k-j)}c_j \nonumber \\
d_k &= \frac{1}{2}\sum_{j=1}^n\left(1-k{b_i}_j^2\right)(2\lambda_j)^{-k} \nonumber \\
\vect{b_i} &= \vect{\Sigma}^{-\frac{1}{2}}(\vect{x}-\vect{a}_i) \nonumber \\
\vect{\lambda} &= eigenvalues(\vect{\Sigma})
\end{align}
where $n=3$ is the number of spatial dimensions and $\vect{\Sigma}$ is the sum of the covariance of the query point and of the nearest neighbor. The probability of collision between a query point and all nearest neighbors is calculated as in \cite{florence2018nanomap}:
\begin{align}
p(\vect{x}) = 1 - \prod_{i=1}^{K}\left[1-p_i \right]
\end{align}

The gradient for this constraint is calculated as:
\begin{align}
    \nabla p(\vect{x}) &= (1 - p(\vect{x}))\sum_{i=1}^{K}\left[\frac{\nabla p_i(\vect{x})}{1-p_i(\vect{x})}\right] \nonumber \\
    \nabla p_i(\vect{x}) &= \sum_{k=0}^{\infty}\left[(-1)^k\nabla c_k\frac{(r+s)^{2^{\frac{n}{2}+k}}}{\Gamma(\frac{n}{2}+k+1)}\right] \nonumber \\
    \nabla c_0 &= c_0\cdot\vect{\Sigma}^{-\frac{1}{2}}\vect{b_i} \nonumber \\
    \nabla c_k &= \frac{1}{k}\sum_{j=0}^{k-1}\left[d_{k-j}\cdot\nabla c_{j}+\nabla d_{k-j}\cdot c_{j}\right] \nonumber \\
    \nabla d_k &= -\frac{1}{2}k{(2\vect{\lambda})^{-k}}^{T}\vect{\Sigma}^{-\frac{1}{2}}\vect{b_i}\
\end{align}

We significantly sped up these probability estimation calculations through the use of bottom up dynamic programming for the d and c parameters for each query. We found that the summing a maximum of 75 terms for $p_i(\vect{x})$ was sufficient.  Additionally, the assumption of a diagonal covariance matrix significantly speeds up the calculation of $\vect{\Sigma}^{-\frac{1}{2}}$ and $\vect{\lambda}$.
\label{sec:approach}

\section{Real-Time Simulation Study}

\subsection{Simulation Setup}

We evaluated the effectiveness of our approach in an unknown map through simulation experiments. Each experiment consisted of 100 trials. A trial was considered successful if the fixed-wing navigated around the hallway without collision.

The dynamics simulator and the controller consist of custom C++ ROS nodes. The depth camera is simulated using Gazebo at 30Hz with a resolution of 128x85, field of view of 87x58 degrees, and a 20 meter sensing range. The dynamics simulator provides the Gazebo simulation with state information for the camera and generates the simulated point cloud data input to the controller. The camera is rigidly attached to the fixed-wing behind the propeller.

The map is a tight hallway with two 90 degree turns, which showcases the aerobatic post-stall maneuvers the system is required to make around buildings in urban environments. Additionally, the 90 degree turns obstruct the camera field of view. 

\begin{figure*}
  \centering
  \subfigure[]{\includegraphics[width=0.3\textwidth]{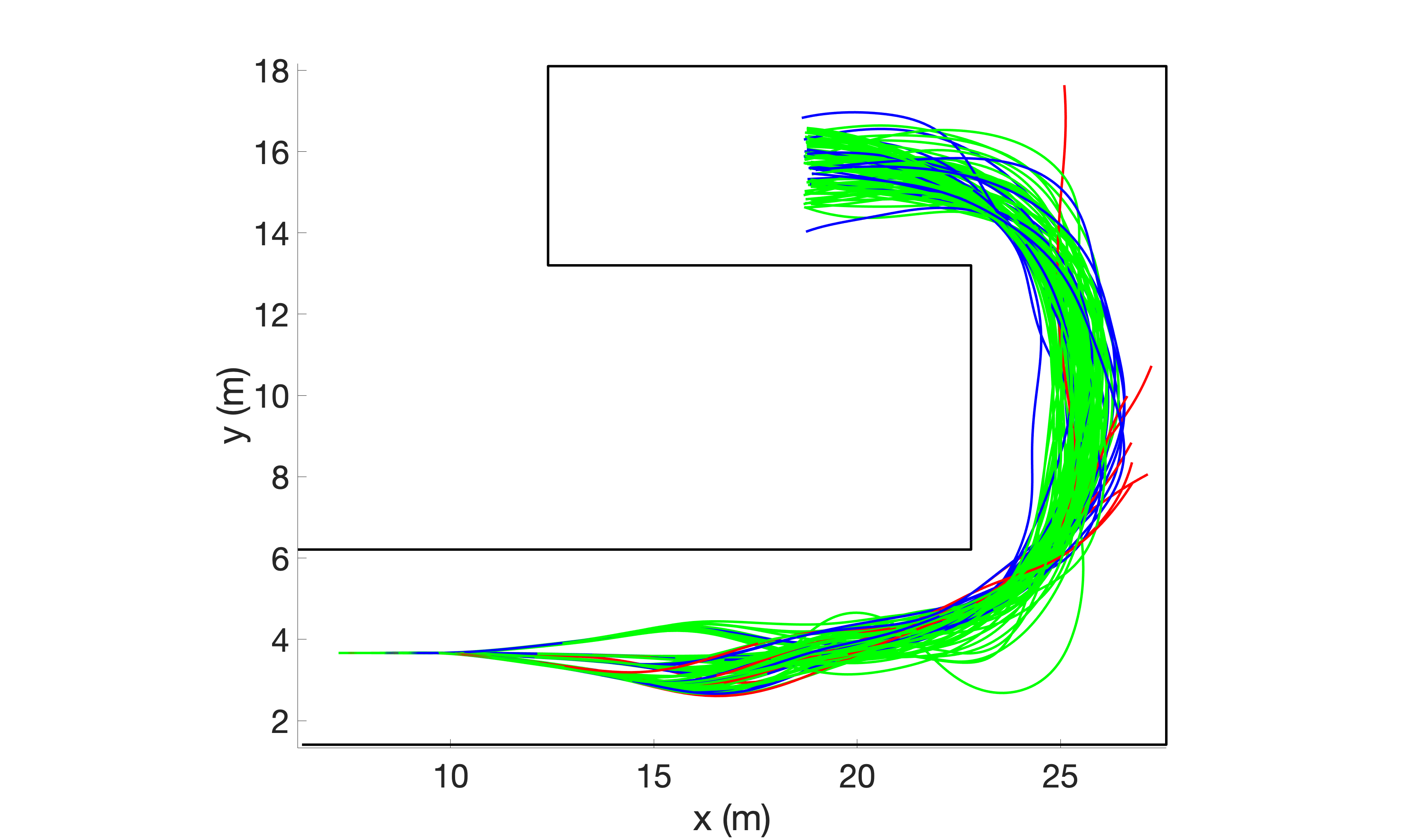}}
  \subfigure[]{\includegraphics[width=0.3\textwidth]{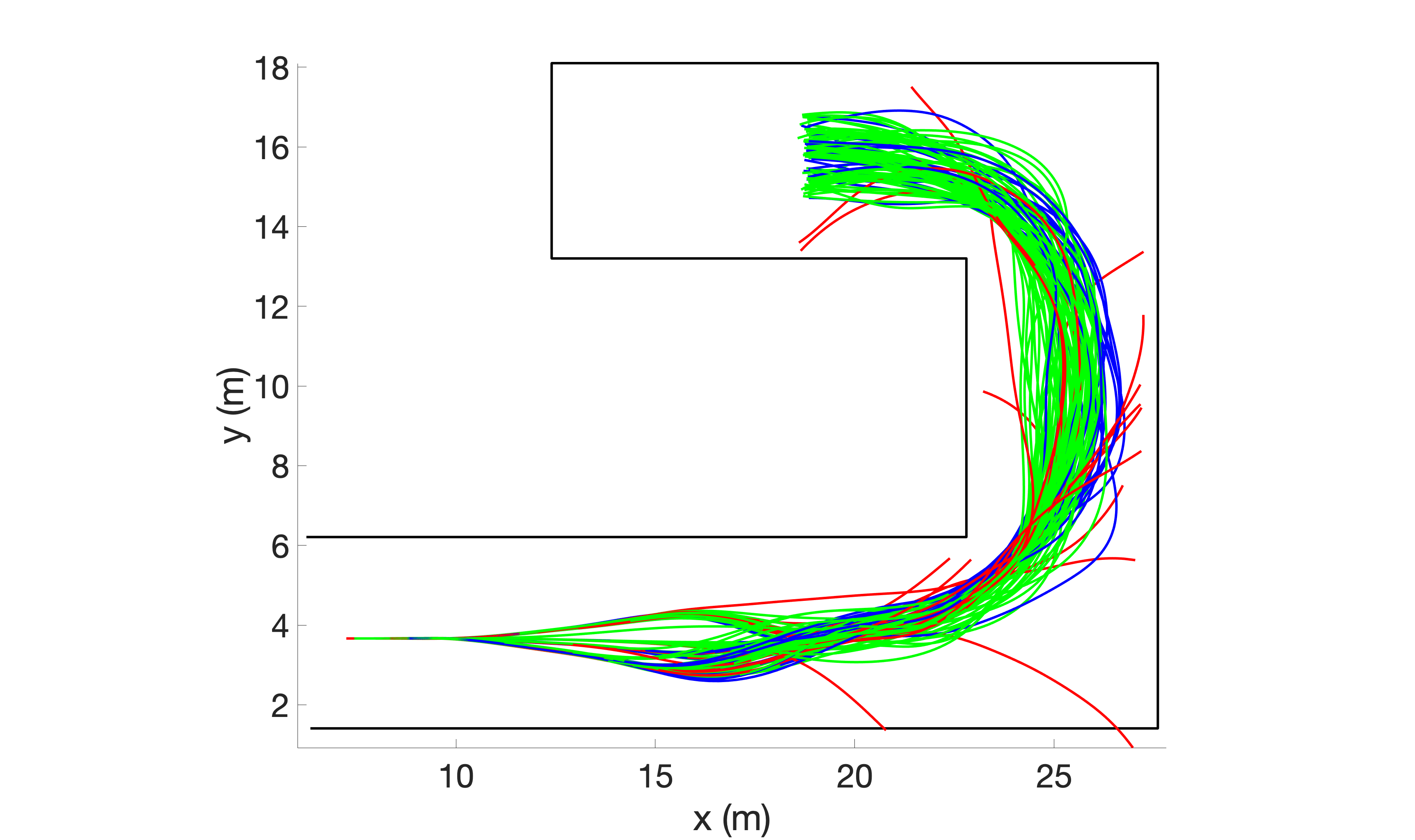}}
  \subfigure[]{\includegraphics[width=0.3\textwidth]{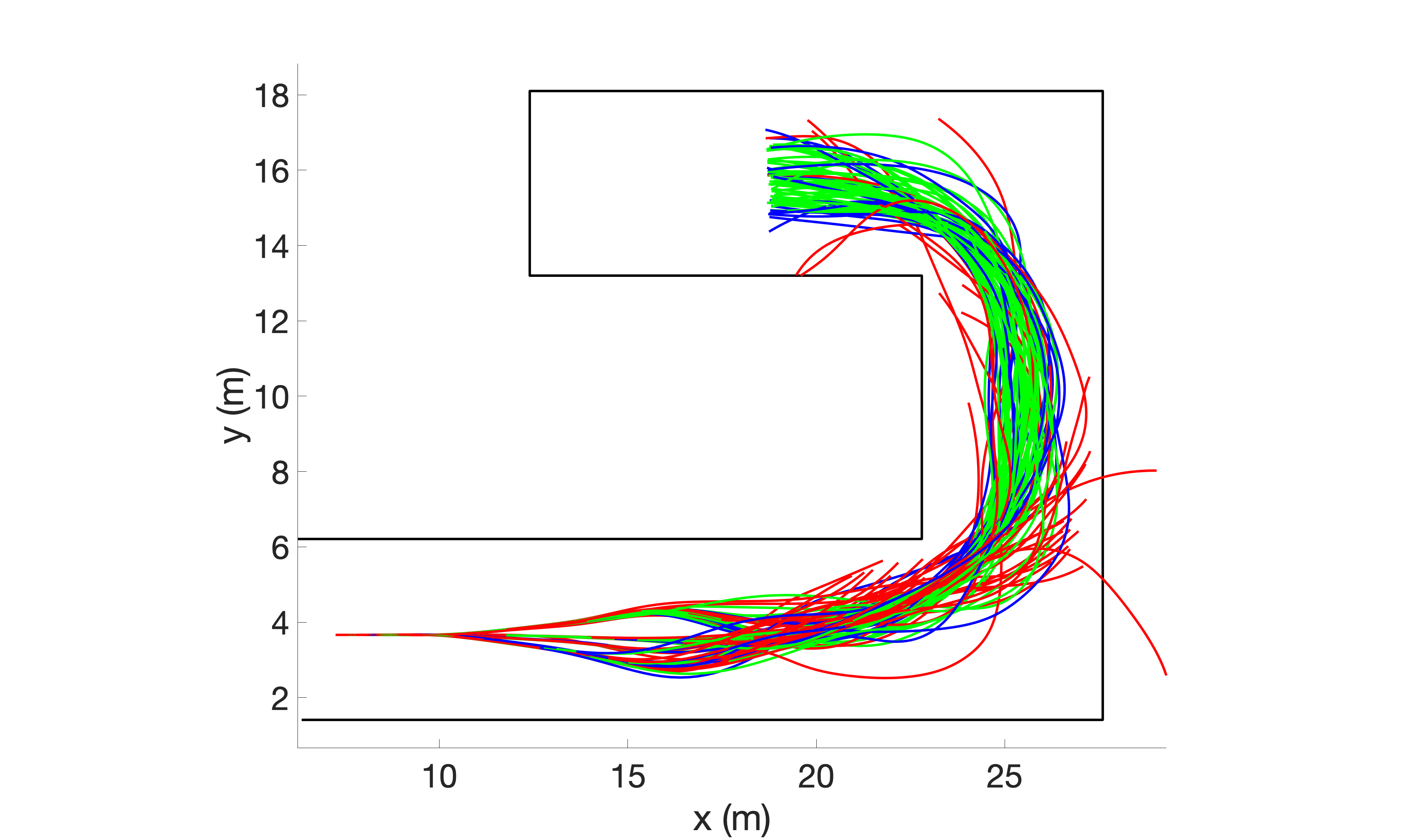}}

  \subfigure[]{\includegraphics[width=0.3\textwidth]{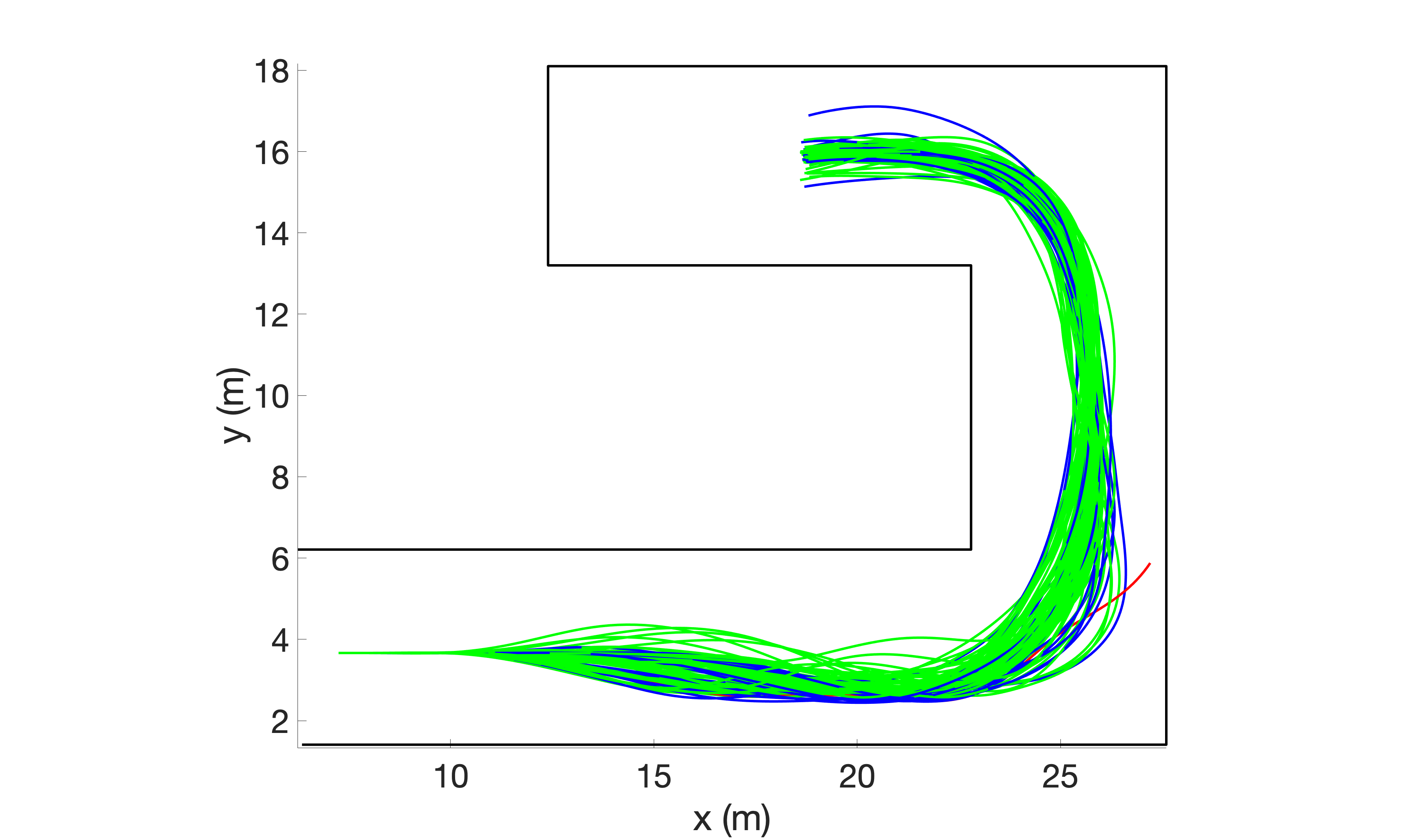}}
  \subfigure[]{\includegraphics[width=0.3\textwidth]{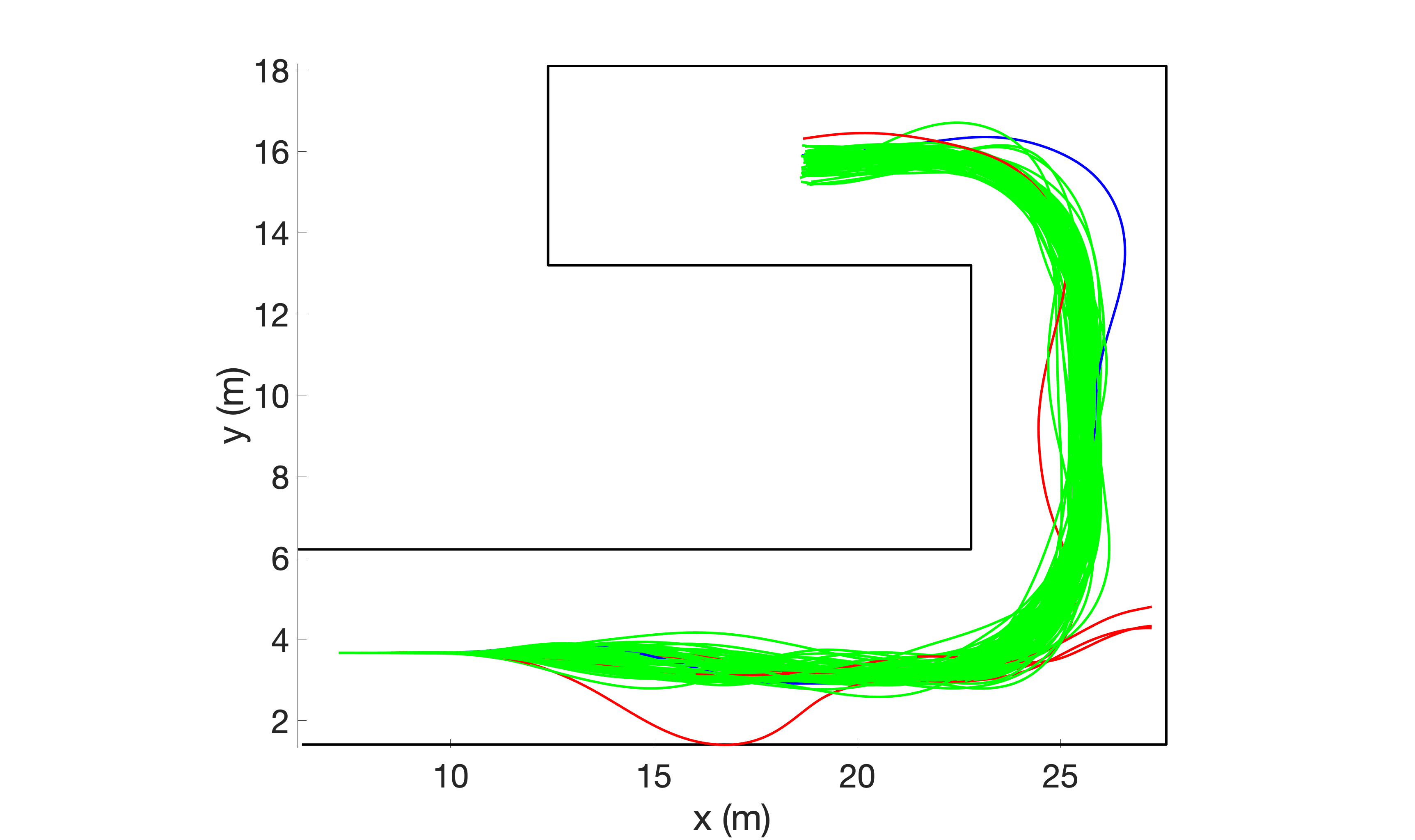}}
  \subfigure[]{\includegraphics[width=0.3\textwidth]{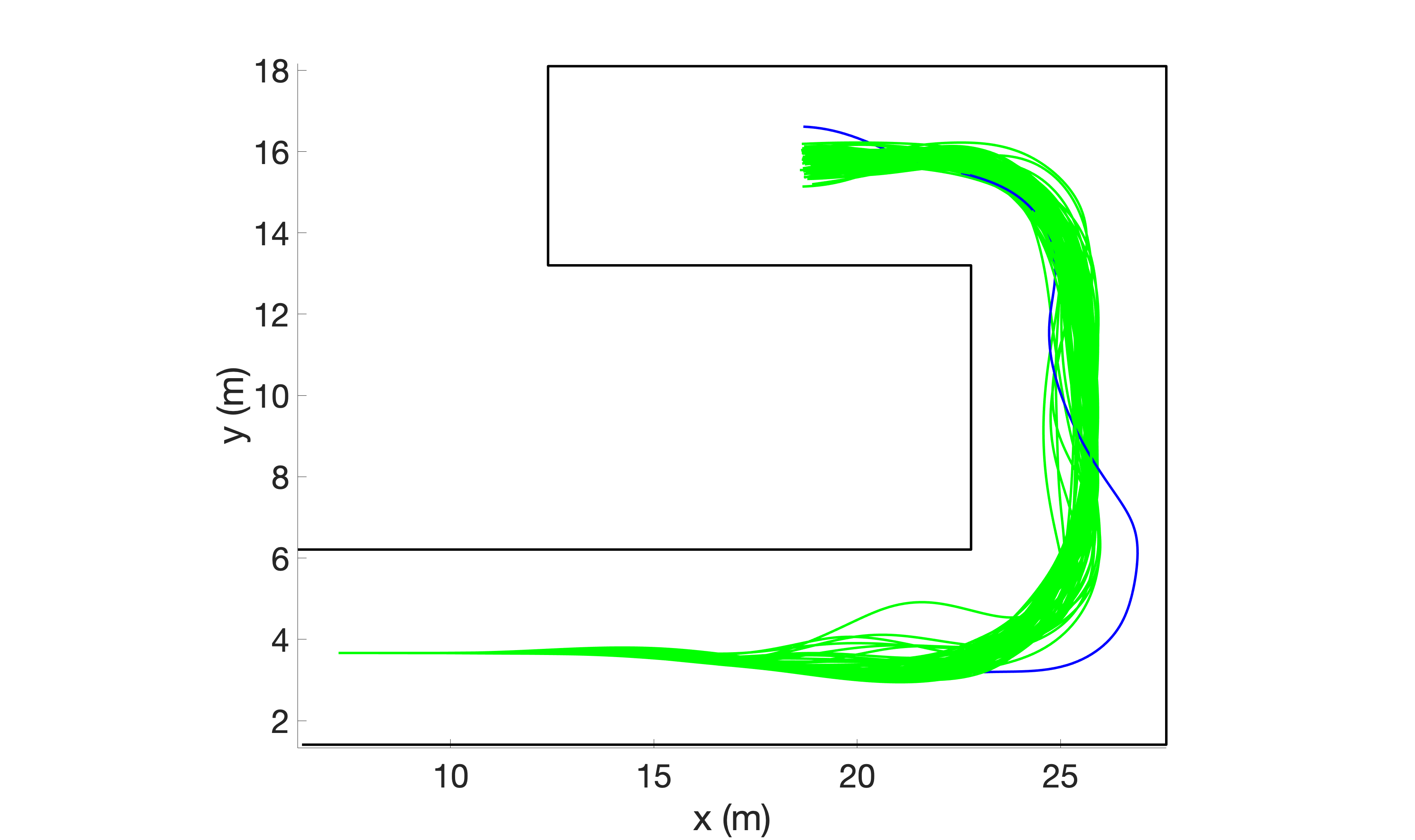}}
  \caption{Successful trials in green, trials without collision but with constraint violation in blue, trials with collisions in red. (a) Trials using full NanoMap history. (b) Only direct transcription restricted to most recent point cloud. (c) RRT and direct transcription restricted to most recent point cloud. (d) Distance to obstacle constraint. (e) Distance to obstacle constraint using standard deviation inflation. (f) Probability of collision constraint. (d), (e), and (f) include measurement and state estimation noise.}
  \label{Fig:sim}
\end{figure*}  

\subsection{Experiment 1: Use of NanoMap History}

The first set of simulation experiments showcases the necessity of reasoning about field-of-view history. The controller was run in simulation without any prior knowledge of the map, but with perfect state information and noiseless depth data. The controller was run using the distance to obstacles constraint with a minimum distance constraint of 1.2 meters. In this experiment, a cost was placed on the final state of the trajectories in lieu of a hard constraint.

NanoMap with a history of 50 point clouds was used. The system was successful in navigation to goal for 91 trials, of which only 15.4\% broke the obstacle constraint, as shown in Figure \ref{Fig:sim}a. 

\begin{figure}[h]
  \centering
  \includegraphics[width=.6\linewidth]{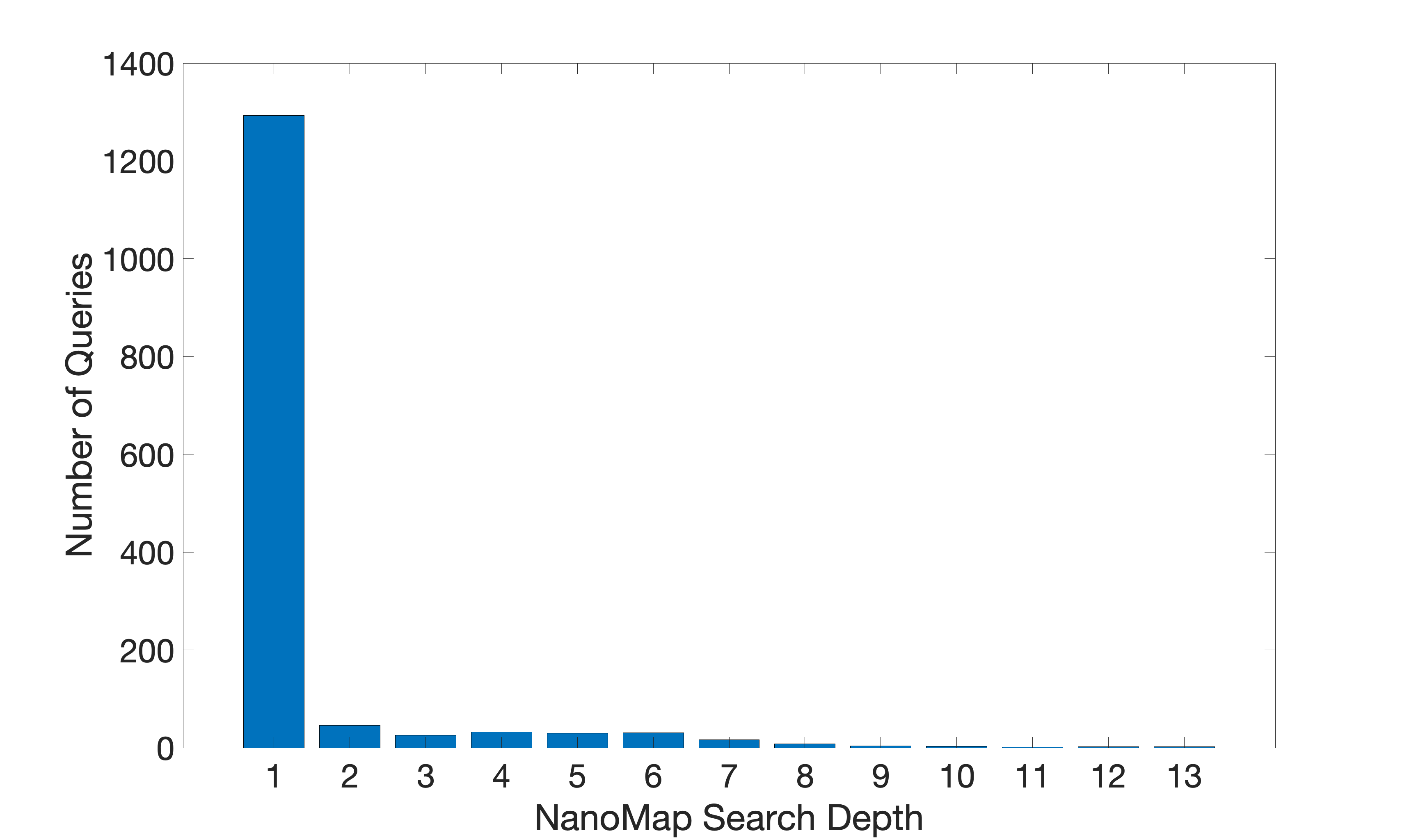}
    \caption{NanoMap search depth queries across all iterations for the trajectory optimizer.}
  \label{Fig:total_search}
  \end{figure}

\begin{figure}[h]
  \centering
  \includegraphics[width=.6\linewidth]{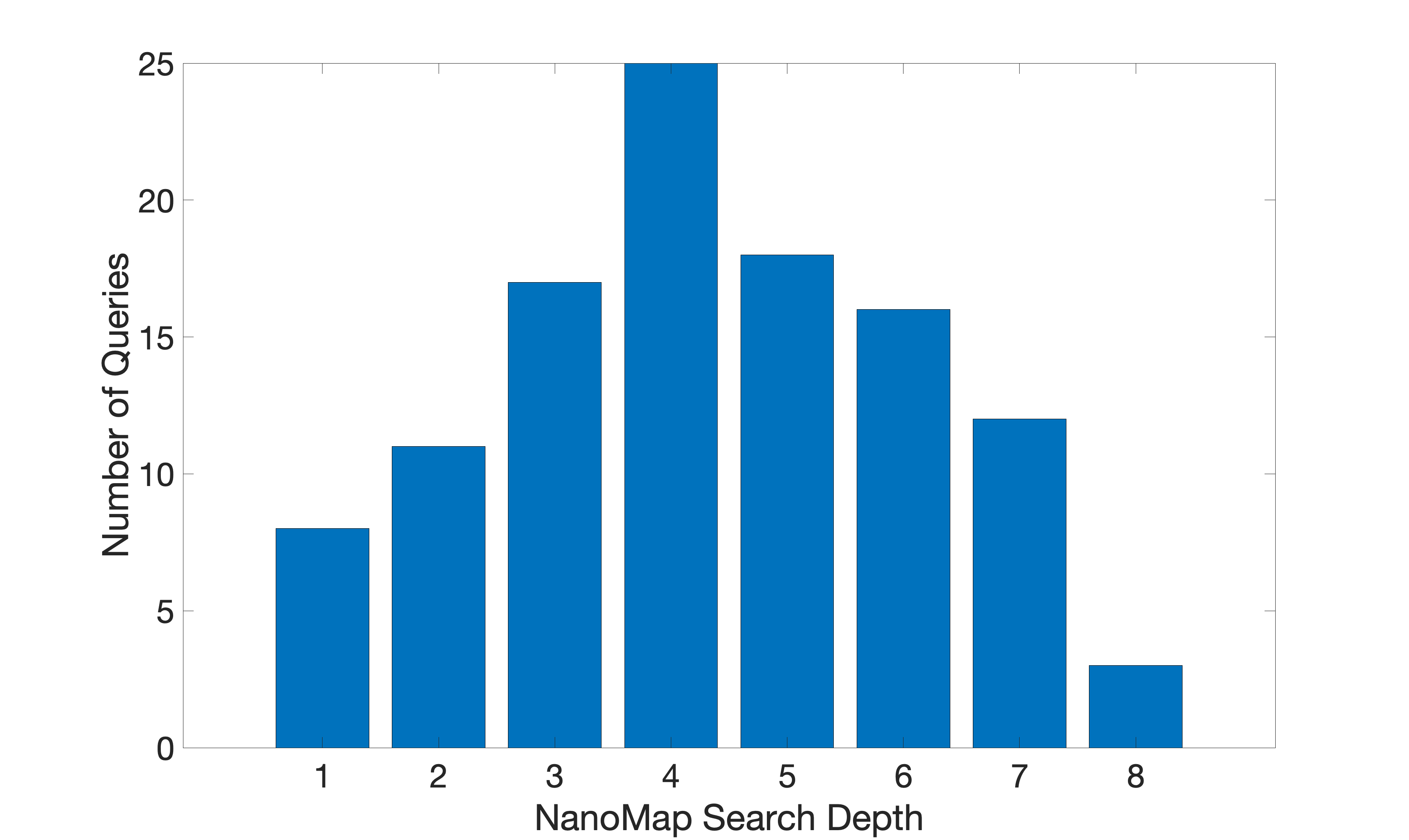}
    \caption{NanoMap search depth queries during optimization of a single trajectory while turning a corner.}
  \label{Fig:turn_search}
\end{figure}

To highlight the trajectory optimizer's use of the NanoMap measurement history in these trials, we tracked the number of queries that occurred at a point cloud history depth for each iteration during the trial. Figure \ref{Fig:total_search} shows the NanoMap search depth during direct transcription across all trajectory generation iterations. Figure \ref{Fig:turn_search} shows the NanoMap search depth during one iteration as the fixed-wing turns a corner. For these data, it is clear that the optimizer makes use of NanoMap's point cloud history as it performs aerobatic maneuvers that require the fixed-wing to quickly turn outside of its field of view.


Next, we performed the same simulation using only the most recent point cloud measurement instead of NanoMap's history. When doing so, the system was only successful 58 times, of which 57.4\% broke the obstacle constraint (Figure \ref{Fig:sim}c). The failure case for the majority of these trials was due to the RRT global path planning failure. 

Thus, to evaluate the effect of using NanoMap on trajectory generation in particular, we allowed the RRT planner to access NanoMap, but restricted the direct transcription trajectory generator to the most recent point cloud. The system was only successful in reaching the goal without collision 80 times, of which 22.5\% broke the obstacle constraint (Figure \ref{Fig:sim}b).

\subsection{Experiment 3: Noise}

The second set of simulation experiments showcases the importance of compensating for sensor uncertainty. Positional noise, $\mathcal{N}(diag(0, 0, 0),\,diag(0.1, 0.1, 0.1))$, was injected into the fixed-wing state to simulate state estimation noise. Noise was also injected into the depth images, $\mathcal{N}(0,\,0.316)$, to simulate depth camera noise. To evaluate the effects of noise only on trajectory generation, a global path to goal is provided in lieu of the RRT planner. 

Three different constraints were evaluated; distance to obstacles, distance to obstacles with standard deviation inflation, and probability of collision. For the probability of collision constraint, we set a max probability of collision of 0.5 and a robot radius of 1.2 meters. For the distance to obstacles with standard deviation inflation constraint, we inflate the collision radius using the standard deviation values. 

The distance to obstacle constraint had 99 successful trials, of which only 17.2\% broke the obstacle constraint (Figure \ref{Fig:sim}d). The distance to obstacles with standard deviation inflation trials had 95 successful trials, of which only 1\% broke the obstacle constraint (Figure \ref{Fig:sim}e). The probability of collision constraint had 100 successful trials, of which only 1\% broke the obstacle constraint (Figure \ref{Fig:sim}f). Thus, we can conclude that explicitly taking the covariance of our state and noise into account benefits planning.

\label{sec:performance}

\section{Hardware Experiments}

\subsection{Experimental Set-up}



\begin{figure}
  \centering
  \includegraphics[width=.9\linewidth]{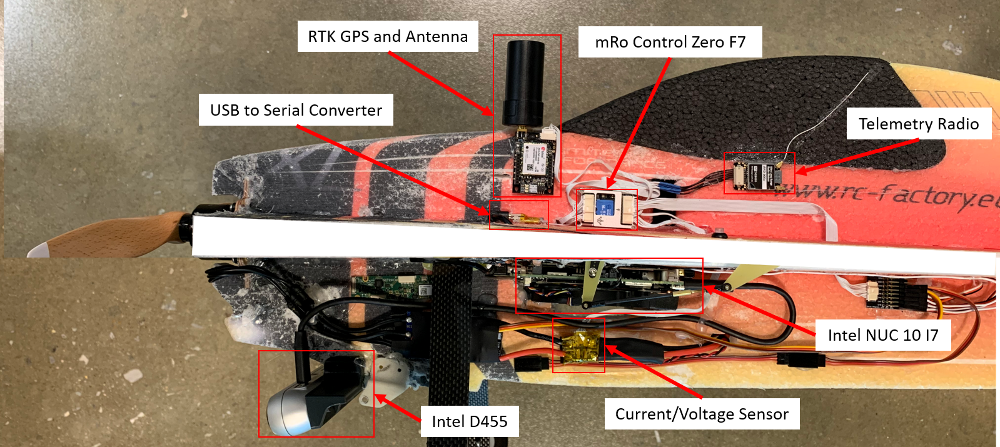}
    \caption{A close-up view of the Edge540XL vehicle and the key components for onboard sensing and control.}
  \label{Fig:labeled_plane}
\end{figure}

To evaluate our method on physical hardware, we used an EPP Edge 540 XL from Twisted Hobbys. State estimation was provided by an mRobotics Control Zero F7 flight controller running Ardupilot firmware. Additional external sensors including a uBlox ZED-F9P RTK GPS receiver and an MS5525 airspeed sensor were used to improve the flight controller's state estimation. An RFD900u telemetry radio was used to monitor the aircraft in flight as well as provide RTK corrections to the GPS receiver. A USB to Serial adapter passed state data from the flight controller to an Intel NUC 10 i7 where our control algorithm ran. An Intel Realsense D455 camera was rigidly attached to the fuselage behind the propeller. Figure \ref{Fig:labeled_plane} shows an image of the plane with labeled components.

The mission for the UAV was to fly towards two waypoints, both of which were obstructed by a large building. This environment showcases both straight, unobstructed zones as well as hard obstacles. Navigation to the waypoints requires the system to make tight turns between the building and trees, which is similar to our hallway simulation environment. A mission was considered successful if the fixed-wing was able to navigate around the building.

\subsection{Simulated Perception}

To evaluate our method using perfect perception data, we first simulated the depth camera data using Gazebo onboard the plane during a physical test flight. To do this, we provided the Gazebo simulation with an accurate mesh of the building and nearby light poles. As the fixed-wing flew around the building, it sent odometry measurements to the simulator. In return, it received noiseless point cloud data given the camera field of view. The fixed-wing used the distance to obstacles with standard deviation inflation with a 3 meter radius, a time horizon $T_H=1s$ and a replanning frequency of 5Hz. 

\begin{figure}
  \centering
  \includegraphics[trim={0 150 0 0},clip,width=.9\linewidth]{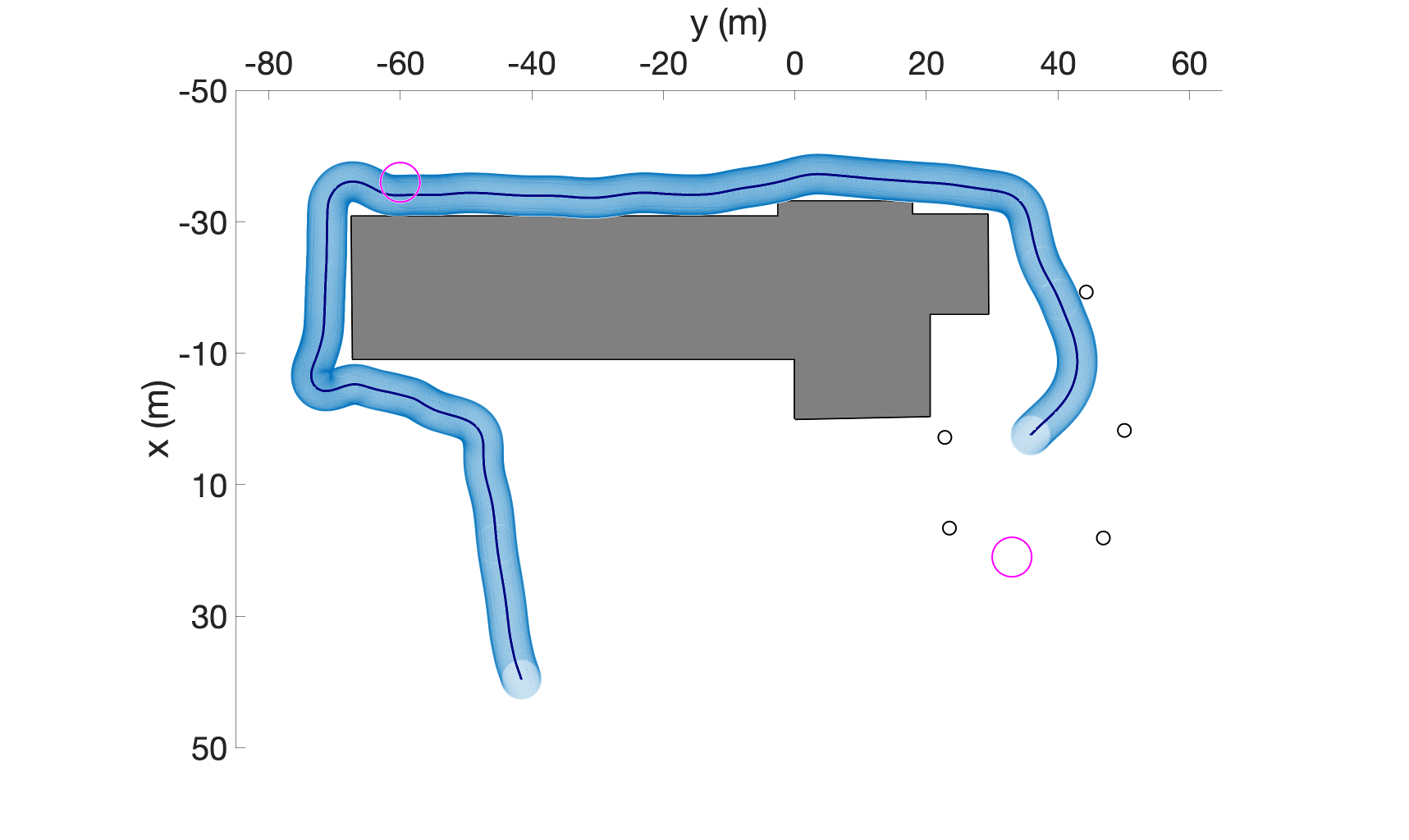}
    \caption{Plot shows the flight path taken by the Edge540XL vehicle around a physical building using a simulated depth camera for navigation. The gray shaded region represents the building and the black circles represent the light poles.}
  \label{Fig:gazebo}
\end{figure}

The fixed-wing was successfully able to navigate around the building to the waypoints using the simulated depth camera data. Figure \ref{Fig:gazebo} shows the fixed-wing's flight path overlaid with the obstacle distance constraint radius.

\subsection{Control Experiment}

After confirming that our method worked on hardware with perfect perception data, we ran the same experiment with the same controller parameters using the depth data from the mounted D455. The depth images from the camera were down sampled by a factor of 10 and converted to point clouds for NanoMap. 

Out of six hardware trials, four successfully rounded the building. In one of these trials, the fixed-wing crashed into a light pole after making it around the building. This crash was caused by the inability of the depth sensor to see the pole until the fixed-wing was too close to react.

\begin{figure}
  \centering
  \includegraphics[trim={0 200 0 0},clip,width=.9\linewidth]{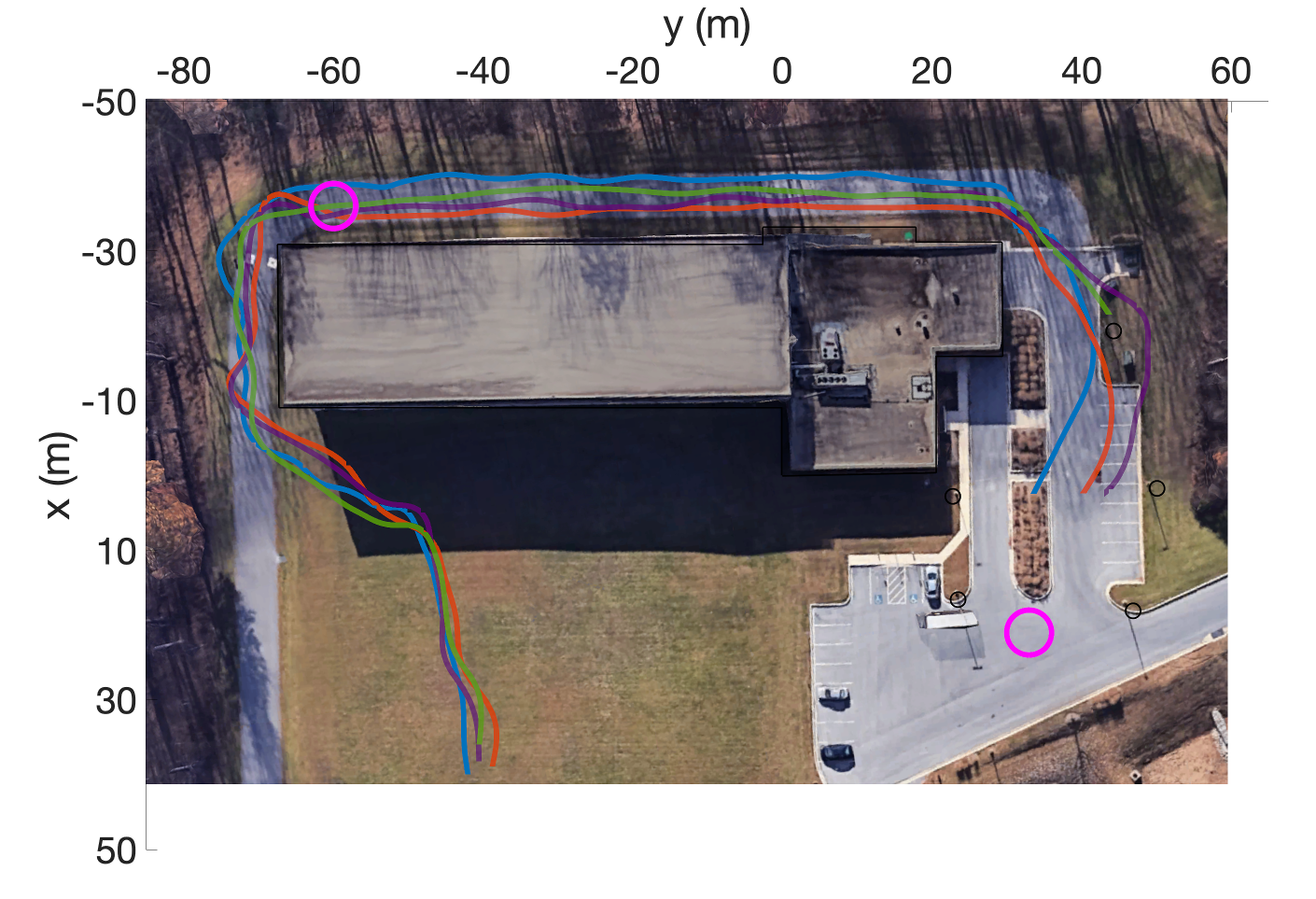}
    \caption{An overhead view of four flight paths taken by the Edge540XL vehicle around a physical building using the onboard RealSense D455 stereo depth camera for navigation. The objective of the controller is to navigate, without collision, to the two pink circle goal regions in sequence.}
  \label{Fig:successful}
  \end{figure}

\begin{figure}
  \centering
  \includegraphics[trim={0 150 0 0},clip,width=.9\linewidth]{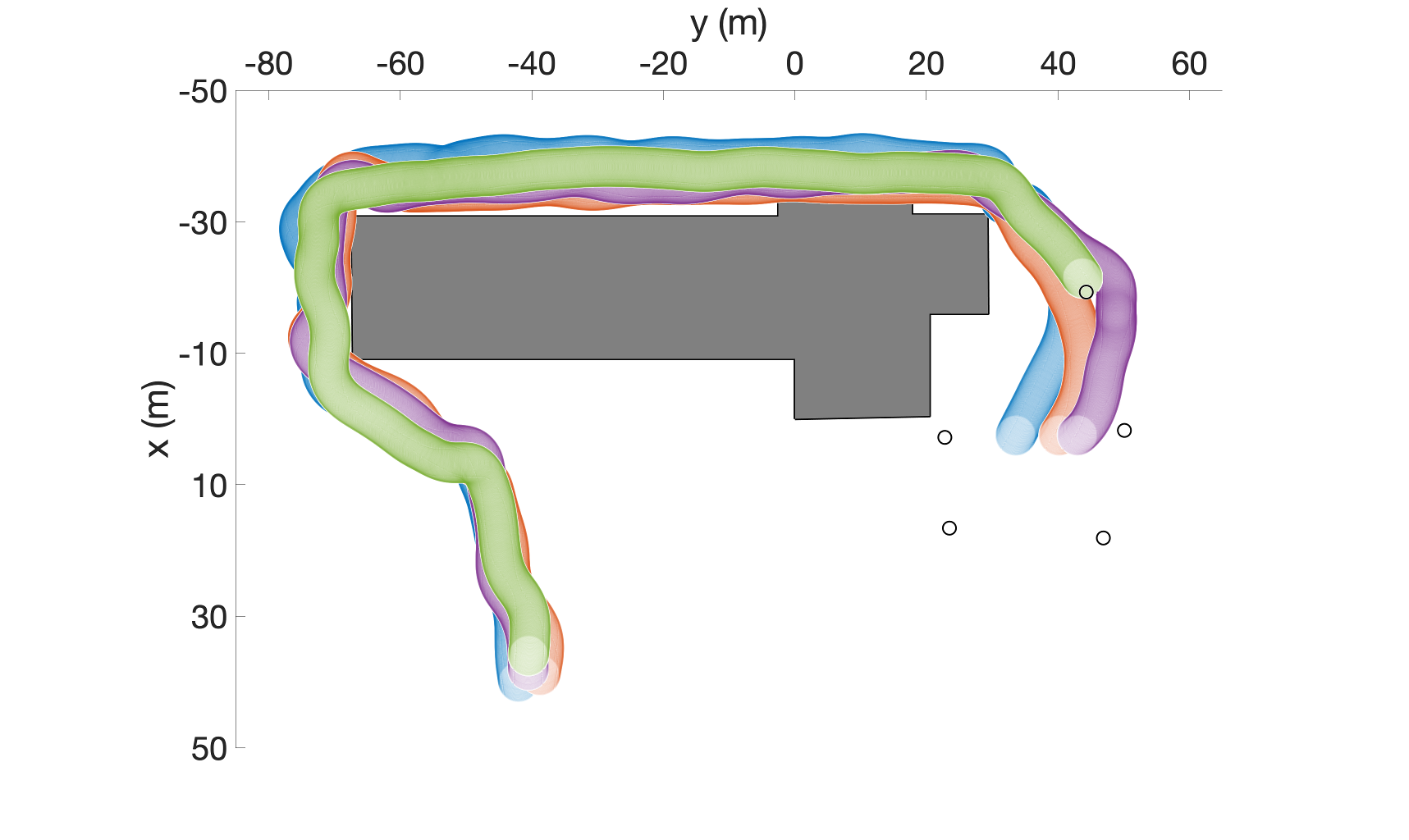}
    \caption{An overhead view of the four flight paths taken by the Edge540XL vehicle with the desired collision radius superimposed to highlight obstacle avoidance performance.}
  \label{Fig:successful_radius}
\end{figure}

\begin{figure}
  \centering
  \includegraphics[width=.70\linewidth]{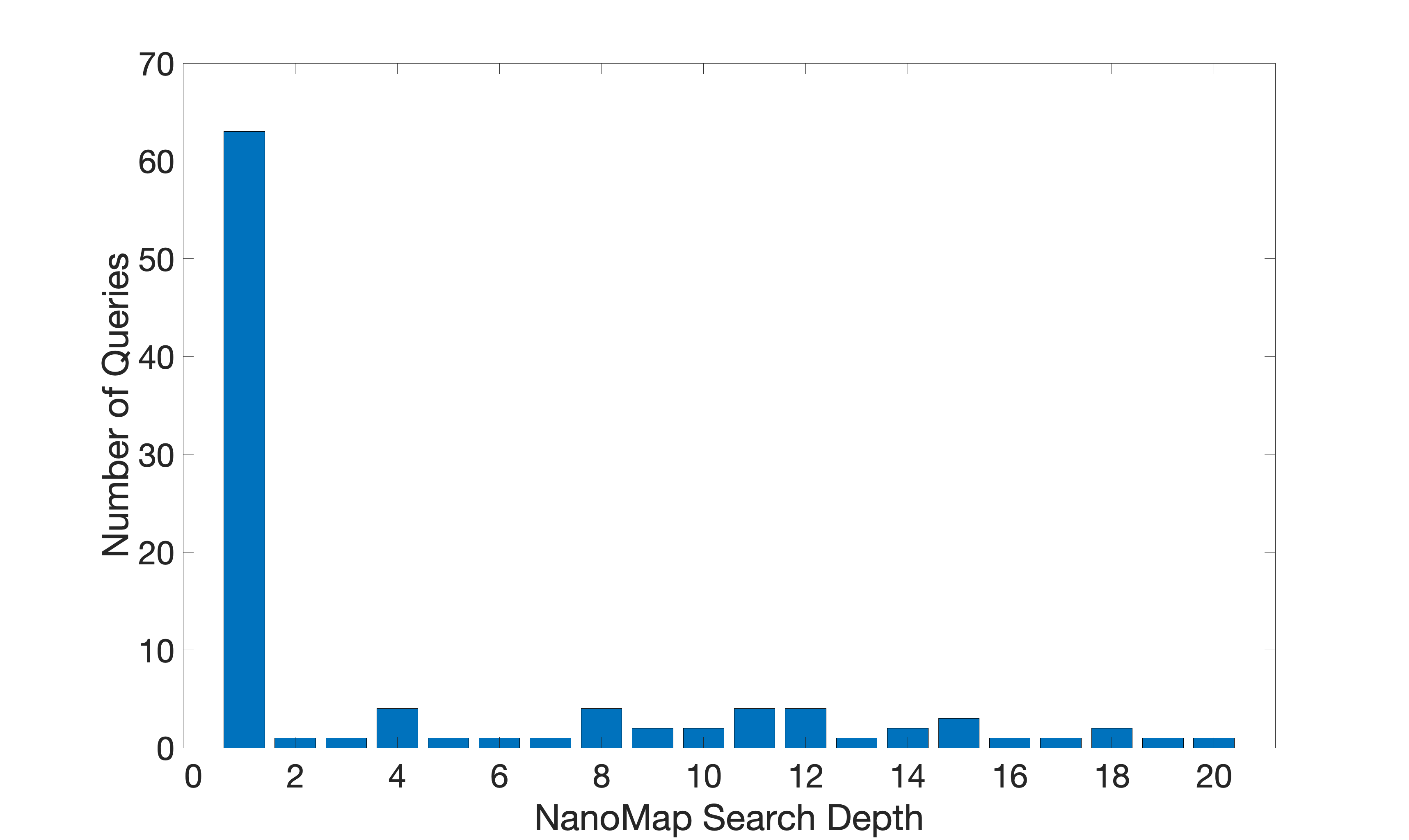}
    \caption{NanoMap search depth queries for a hardware post-stall turn.}
  \label{Fig:library_turn_search}
\end{figure}

The paths of the successful trials are shown in Figure \ref{Fig:successful}, and the obstacle distance constraint radii are overlaid in Figure \ref{Fig:successful_radius}. Additionally, the search depths of NanoMap queries during a turn are shown in Figure \ref{Fig:library_turn_search}.



\begin{figure}
  \centering
  \includegraphics[trim={0 800 0 0},clip,width=.9\linewidth]{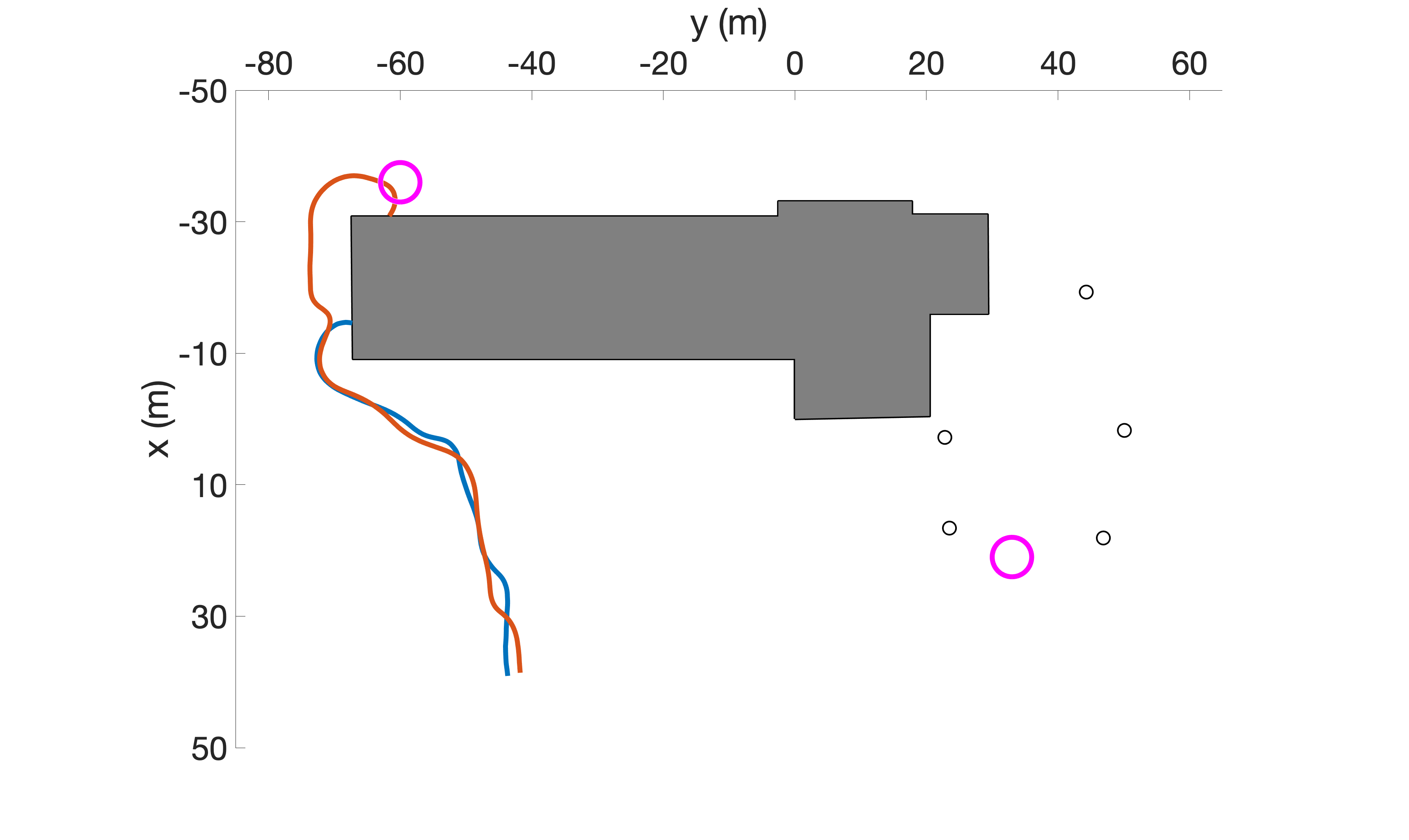}
    \caption{Plot shows the flight paths of the failed hardware trials. }
  \label{Fig:failed}
\end{figure}

Two of the hardware trials failed, crashing into the building shortly after executing a turning maneuver (see Figure \ref{Fig:failed}). In both of these failure cases, the fixed-wing turned towards the building very late after rounding the corner. When this happens, the depth camera does not observe the part of the wall after the corner, RRT paths are generated through the apparent gap, and the vehicle crashes.

\label{sec:experiments}

\section{Discussion}

In this work, we outlined a controller capable of enabling agile fixed-wing flight in unknown environments. In simulation, we showed the ability to navigate a tight hallway environment with a fixed-wing UAV. We also demonstrated the ability of a fixed-wing to avoid and navigate around a building without prior knowledge of its environment. In the future, we plan to explore improving system robustness by more tightly coupling perception and control. For instance, by augmenting the optimizer's cost function, we may be able to reduce failure cases by rewarding information gathering during aggressive maneuvers. We may also be able to improve performance by using generative adversarial networks, as was done in \cite{katyal2020high}, to predict and reason about the environment beyond the field of view.

\label{sec:discussion}

\section{Acknowledgement}
This material is based upon work supported by the DARPA OFFSET program. The views, opinions and/or findings expressed are those of the author and should not be interpreted as representing the official views or policies of the Department of Defense or the U.S. Government.
\bibliographystyle{IEEEtran}

\pagebreak

\bibliography{references}

\end{document}